# VocalTweets: Investigating Social Media Offensive Language Among Nigerian Musicians

**Sunday Oluyele[1], Juwon Akingbade[1], Victor Akinode[1]**
[1]Department of Computer Engineering, Federal University Oye Ekiti, Nigeria
{sunday.oluyele.2826, juwon.akingbade.1011}@fuoye.edu.ng
victorakinode@gmail.com

## Abstract

Musicians frequently use social media to express their opinions, but they often convey different messages in their music compared to their posts online. Some utilize these platforms to abuse their colleagues, while others use it to show support for political candidates or engage in activism, as seen during the #EndSars protest. There are extensive research done on offensive language detection on social media, the usage of offensive language by musicians has received limited attention. In this study, we introduce VocalTweets, a code-switched and multilingual dataset comprising tweets from 12 prominent Nigerian musicians, labeled with a binary classification method as Normal or Offensive. We trained a model using HuggingFace's base-Twitter-RoBERTa, achieving an F1 score of 74.5. Additionally, we conducted cross-corpus experiments with the OLID dataset to evaluate the generalizability of our dataset.

## 1. Introduction

Social media platforms like Twitter (X), Facebook, and the likes have exerted extreme influence on how individuals, including public figures, interact and communicate with their audience, which can yield positive or negative effects. This platform has created opportunities for people in various parts of the world to express and share their thoughts instantaneously and extensively. Public figures or so-called celebrities, like musicians with large follower bases on social media platforms, often play a significant role in shaping public opinion and discourse (Jin & Phua, 2014). Their capabilities and influence, most times can be a powerful tool for spreading information and engaging with people on a global scale; studies have shown that when celebrities engage in social or political issues, their messages often reach a wide audience and influence public attitudes (Marshall, 2014). However, there has been an increased pervasiveness of offensive language on social media, such as Twitter (X), which poses a significant threat to healthy online interactions (Mutanga et al. 2020).

Before, offensive language detection primarily relied on traditional approaches like using rule-based systems and classical machine learning techniques. Keyword detection and lexicon-based approaches were used in early methods, where offensive language was identified by predefined lists of offensive words. With the simplicity of these methods, yet they are limited in effectiveness as they find it difficult to account for context, sarcasm or evolving slang. For example, keyword-based systems would flag any occurrence of a specific derogatory term, which often led to false positives when the term was used in non-offensive contexts (Warner & Hirschberg, 2012). Likewise, lexicon-based methods which rely on dictionaries of hate-related words were also ineffective, as they lacked the effectiveness to adapt to different ways mixed languages are being used on social media. With evolution, there was the introduction of machine learning algorithms, such as Support Vector Machines (SVM) and Naive Bayes classifiers to improve detection accuracy. Feature engineering techniques like Term Frequency-Inverse Document Frequency (TF-IDF) and n-grams to represent text were used with these models. While these models gave improved performance over rule-based systems, there are still issues in detecting hate speech in the form of indirect or implicit hate, due to their reliance on manual feature extraction and lack of deep contextual understanding (Davidson et al., 2017).

Offensive language is characterized by sharp, rude content that goes beyond mere negative reactions (Touahri & Mazroui, 2022). It can include vulgar content, insults, or attacks against individuals or groups (Mubarak et al., 2020). Thi has been shown to have severe psychological impacts, contributing to stress, anxiety, depression, and even PTSD in victims and witnesses. The constant exposure to derogatory or harmful content online can also lead to emotional desensitization and social isolation (Wachs et al., 2020).

In regions with varied linguistic settings and diverse social structures like Africa, the matter of hate speech becomes even more complex because the difference in various languages and dialects, combined with distinctive cultural contexts, makes detecting and addressing hate speech particularly challenging in these areas. Researchers often face the challenge of creating comprehensive datasets to effectively detect hate speech on social media. These datasets must account for current trends in online communication, including emoticons and emojis, hashtags, slang and colloquialisms, and linguistic contractions. An extensive dataset containing these elements is important for developing accurate hate speech detection systems that can keep pace with evolving online communication styles (Mody et al., 2023). With the significant influence of social media particularly amidst public figures, it is important to understand the extent and nature of offensive language within these interactions. Since traditional methods of detecting offensive language have shown limitations in diverse linguistic environments, advancements in machine learning offer new opportunities. While significant research has been explored in the Western world in the area of offensive language detection, there are gaps that still need to be filled in Africa. African musicians often use multilingual language in their communication, making it difficult to analyze their social media interactions using existing Natural Language Processing (NLP) models. In this work, the focus is on Nigerian musicians because this set of people tends to engage in different diverse topics, and sometimes there are quarrels among these musicians; this is evident mostly between Wizkid and Davido. Some of them also make use of this platform to criticize individuals, organizations, and the government, which sometimes also leads to the abusive use of speech. Research shows that the spread of offensive language online has increased dramatically in recent years, posing significant social and political risks (Citron & Norton, 2011). Therefore, this study aims to address how these prominent Nigerian musicians, such as Davido, Burna Boy, Wizkid, Don Jazzy, Tiwa Savage, Peter Psquare, Vector, Buju, Erigga, Falz, Teni, and Tems, interact on Twitter (X) by examining their offensive language patterns. This will be done by creating a comprehensive dataset that captures how Nigerians make use of speech on social media platforms, investigating the prevalence of offensive language among these musicians, and advancing natural language resources for African languages. This will thus contribute to developing datasets for building Natural Language models in Africa to enhance support for offensive language detection across diverse linguistic landscapes.

The rest of this paper is structured as follows: The next section discusses some selected and relevant related works. In Section 3, details of the approach involved in data collection and model training are provided. Results and Findings are presented in Section 4. The final Section 5 concludes the paper with highlights of the major findings and recommendations for future work.

A recent study by Ilevbare et. al., (2024) introduced EkoHate, a dataset focusing on code-switched political discussions on Nigerian Twitter(X). 3,398 tweets were collected and annotated from the 2023 Lagos gubernatorial election discussion on Twitter(X), featuring code-switching between English, Yoruba, and Nigerian Pidgin. Both binary and fine-grained four-label annotation schemes were used to categorize tweets as normal, abusive, hateful, or expressing contempt. Baseline experiments were conducted using Twitter-RoBERTa, achieving strong performance on the binary classification task. Their cross-corpus transfer experiments with existing datasets demonstrated EkoHate's ability to generalize to political discussions in other regions. Likewise, their error analysis highlighted challenges in distinguishing between abusive and hateful content, particularly code-switched tweets. This work addresses a significant gap in hate speech detection research for African languages and political contexts, providing a valuable resource for studying online discourse in Nigeria. Yuan and Rizoiu (2024) proposed a Multi-Task Learning (MTL) framework that improves cross-dataset generalization in hate speech detection models. This introduces a novel MTL approach that simultaneously trains on multiple hate speech datasets. This method helps capture different definitions of hate speech and remarkably enhances performance across unseen datasets compared to state-of-the-art models. Also, the PubFigs dataset was introduced, a new dataset focusing on American public political figures, further underscoring the model's ability to generalize across different hate speech domains. By employing MTL and transfer learning techniques, the study demonstrates that hate speech detection can benefit from learning generalized representations that account for different labeling practices and definitions. Egode et al. (2023) explored hate speech detection on Twitter using NLP and deep learning techniques. Naive Bayes, SimpleRNN,

LSTM, and GRU models were compared on a dataset of 31,962 labeled tweets, in which their best performing LSTM model achieved an accuracy of 0.950, precision of 0.633 and recall of 0.674. While demonstrating the viability and potential of deep learning for hate speech detection, the limitations of limited training data and class imbalance were pointed out. The authors suggested future work should investigate BERT and ensemble methods to enhance classification accuracy.

Badjatiya et al. (2017) used deep learning models, including CNNs, LSTMs, and FastText, to detect hate speech on Twitter. It was found that these models outstandingly outperformed traditional methods like TF-IDF and character n-grams, in which the best results were obtained from LSTM embeddings combined with Gradient Boosted Decision Trees (GBDT), achieving an F1 score of 0.93. This shows how deep learning can improve hate speech classification, especially task-specific embeddings. Gambäck and Sikdar (2017) also suggested using Convolutional Neural Networks (CNNs) for hate speech classification on Twitter data. In their model, tweets were classified into four classes which are racism, sexism, and both—non-hate speech. Various feature representations were tried (char 4-grams, word2vec word vectors, and a combination of the two). Word2vec embeddings produced the highest performance of all models, with an F1 score reaching 78.3%. This work also demonstrates that CNNs can be applied to enhance hate speech detection; word embeddings are invaluable in extracting semantic meaning from text. Okechukwu et al. (2023) also leveraged TF-IDF and a majority voting ensemble incorporating support vector machines, k-nearest neighbors, and logistic regression models in hate speech detection systems. This approach achieved remarkably high accuracy of 95% and an F1-score of 0.95, outperforming previous methods like Gambäck and Sikdar (2017) with an F1-score 0.78 and Badjatiya et al. (2017) with an F1-score of 0.93. This work highlights the effectiveness of ensemble learning by meaningfully improving the identification of online hate speech across social media platforms by reducing individual model biases and variances.

In a review work by Jahan and Oussalah (2023), several natural language processing and deep learning methods, including CNN, LSTM, and BERT, have been used for hate speech detection, which has shown potential but still face challenges due to the complex nature of language and the variety of datasets. It was suggested that future improvements require more comprehensive and multilingual datasets to enhance the accuracy and generalizability of hate speech detection models. According to Chao et al. (2024), during the COVID-19 period, offensive and particularly hateful language was rampant on social media. They developed a two-tier detection model that uses advanced machine techniques, including Large Language Models (LLMs) and latent Dirichlet Allocation (LDA). The model focuses on detecting and transforming hate speech within the Chinese language, achieving an accuracy rate of 94.42% in the first phase and 81.48% in the second. By integrating the techniques of generative AI, the work proposes a rephrasing framework that reduces the hostility of online statements while retaining their original context. Similarly, Khanday et al. (2022) also addressed the issue of hate speech detection during the COVID-19 pandemic using machine learning and ensemble methods but their primary focus was on Twitter. The model used hybrid feature engineering (TF/IDF, Bag of Words, and tweet length) and applied classifiers like Decision Trees and Stochastic Gradient Boosting. Unlike Chao et al., whose work solely focused on transforming hate speech using generative AI, Khanday et al. focused purely on classification achieving an accuracy of 98.04% with Stochastic Gradient Boosting. In detecting hate speech on social media, Ibrahim et al. (2024) used social sensing, capitalizing on big data from platforms like Twitter and Facebook. The limitations of traditional rule-based and basic machine-learning approaches in handling the contextual and evolving nature of hate speech were addressed in their work. Natural Language Processing (NLP) and Machine Learning (ML) techniques were employed to improve detecting accuracy, and models like Support Vector Machines (SVM) and Random Forest were tested. The SVM model outperformed others, achieving an accuracy of 86%.

## 2. Methodology

### The VocalTweet dataset

VocalTweets is a dataset comprising tweets from 12 prominent Nigerian musicians: Falz, Wizkid, Davido, Vector, Erigga, Teni, Tiwa Savage, Peter of P-Square, Buju, Burna Boy, Don Jazzy, and Tems. These artists are influential in the Nigerian musical scene and actively engage with their fan base through social media interactions via X (formerly Twitter). VocalTweets captures their online expressions, which goes beyond their music lyrics but provides a sneak peek into their daily lives, personal opinions, and social activism.

### Data Collection

As mentioned, the dataset comprises 4,677 tweets collected from the musicians' official X accounts. Figure 1 shows the distribution of tweets per musician; VocalTweets has 775 tweets from Davido, who has the highest number, followed by Don Jazzy with 661 tweets, Falz and Tiwa have the lowest number of tweets with 187 and 145, respectively.

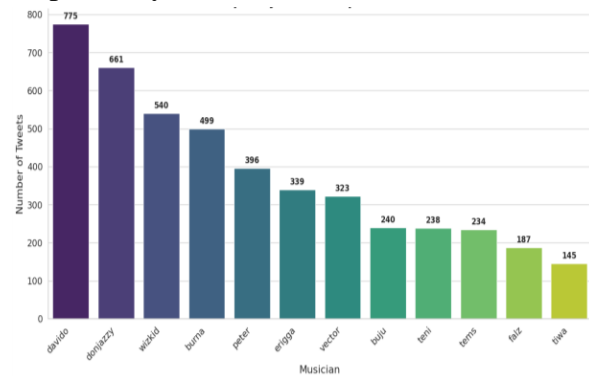

Figure 1: Frequency of tweets per musician

These tweets were manually collected by copying and pasting directly into a Google Sheet. To ensure the dataset's relevance and focus, we intentionally omitted promotional tweets— primarily advertising music releases and concerts—during the collection process.

### Data Annotation

The annotation of the dataset was done using LabelStudio, which is a data labeling tool. We engaged two undergraduate annotators to categorize each tweet into one of the two classes:

- Normal (N): Tweets that do not contain any offensive language.
- Offensive (O): Tweets with offensive sentiments.

Instructions were given to the annotators to ensure unbiased classification to maintain objectivity and consistency. The inter-annotator agreement was measured using Fleiss' Kappa, with a score of 0.76, indicating a good level of agreement between the annotators. In cases with inconsistencies, an objective resolution was resolved, and a final sentiment was allocated to such tweets. After annotation, we observe that many of the annotated tweets are in the Normal category, while others are offensive. Also, more of the tweets are in English, while others are in other languages. An overview of this is visualized in figure 2.

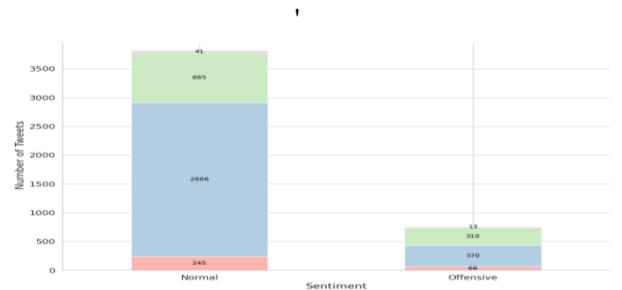

Figure 2: Sentiment and language distribution in VocalTweets

**Class distribution**

The dataset exhibits a notable imbalance in class distribution as there are 3817 (83.4%) normal tweets and 759 (16.6%) offensive tweets, as shown in Figure 3. A closer look at the dataset in figure 4 reveals that specific musicians exhibit higher tendencies of offensive language in their tweets; for instance, 43% of Vector's tweets are offensive; taking a closer look at these tweets, most of them are posted during the Nigerian #EndSARS protest - this was a mass mobilization against the police organization, seeking greater accountability in Nigeria (Ojedokun et al., 2021). In contrast, Don Jazzy has the highest number of normal tweets, with 96.8% of his tweets in the normal category, while only 3.2% are offensive.

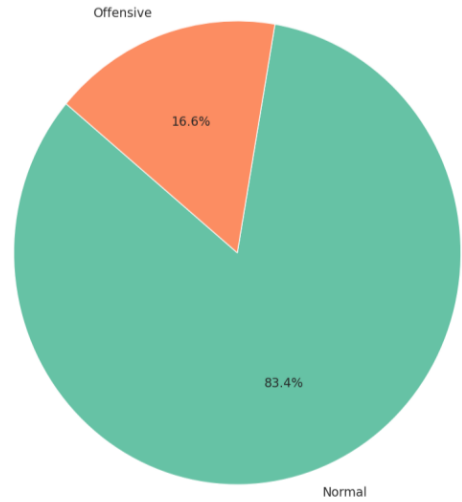

Figure 3: Overall sentiment distribution

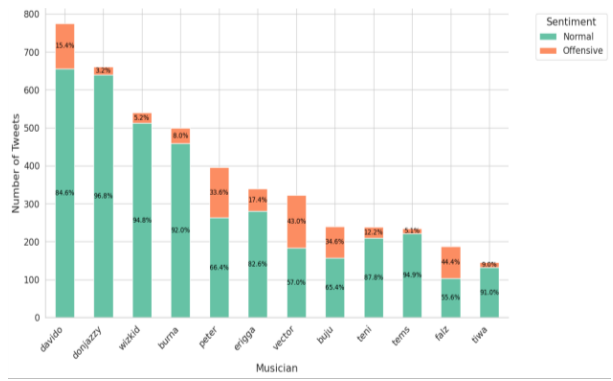

Figure 4: Sentiment distribution of tweets per musician

**Language Distribution**

VocalTweets is characterized by its Multilingual and code-switching nature, which reflects the diverse language usage in Nigeria. Figure 5 reveals that 3,036 (66.3%) tweets are in English, 1,175 (25.7%) are in Nigerian Pidgin (or Naija), 311 (6.8%) are code-switched, and only 54(1.2%) tweets are in Yoruba.

Looking more closely, we observed, as seen in Figure 6, that one of the musicians, named Erigga, predominantly tweets in Nigerian Pidgin, with approximately 60% of his tweets in the language, and Vector also has 42.9% of his tweets in Nigerian Pidgin. Davido has the highest usage of Code-Switching with 12.4% of his tweets switches between Yoruba and English. An example of such a tweet is: *"Orie ti daru that yo bitch ass was crying screaming bout they intimidating ur family u a real bitch for life!!!"*. Teni has the highest usage of Yoruba with about 5% of her tweets in Yoruba.

The dataset possesses noisy characteristics, a common characteristic of social media communications. This is compounded by the fact that these musicians tweet with alphanumeric characters and many emojis.

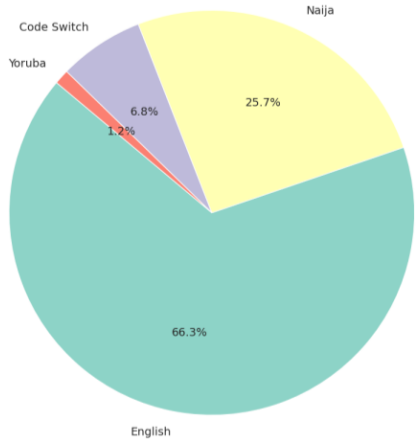

Figure 5: Overall language distribution

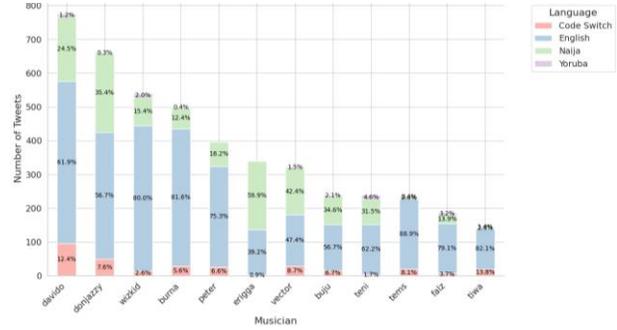

Figure 6: Language distribution of tweets per musician

**Data Splitting**

To facilitate effective training and evaluation in the creation of the model, the dataset was split into training and validation subsets using an 80-20 ratio, with 80%(approximately 3,742 tweets) of the data being the training set and 20% (approximately 934 tweets) of the data being the validation set. This split is necessary to ensure that models trained on the VocalTweets dataset can be evaluated for performance and generalization.

**Experimental Design**

This work uses a binary classification framework to categorize tweets into two classes: Normal and Offensive. The objective is to create a model that can accurately detect non-offensive and offensive language used on social media by Nigerian musicians.

To evaluate the generalizability of our dataset, we conducted cross-corpus transfer experiments using VocalTweets and the Offensive Language Identification Dataset (OLID), which comprises annotated tweets categorized into normal and offensive. We adapted OLID to fit our binary classification framework and conducted cross-corpus transfer experiments on both datasets. For each classification task, in-domain (VocalTweets on VocalTweets and OLID on OLID) and cross-domain (VocalTweets on

OLID and vice versa), we conducted five independent runs with different random seeds to ensure the reliability of our results.

**Model and Training**

We used the Twitter-roberta-base model - a transformer-based architecture fine-tuned specifically for Twitter data by HuggingFace, to perform classification tasks like our study. The models are trained under the following parameters:

- Learning rate: 2e-5
- Batch size: 16 per device
- Epoch: 10
- Weight decay: 0.01
- Logging steps: 50
- Metric for best model: eval_Macro F1

To assess the model's performance, we reported both label-wise F1 scores and the overall Macro F1 Score. The label-wise F1 scores give us the model's ability to correctly classify each class (Normal and offensive). In contrast, the Macro F1 Score gives an idea of the aggregated performance measure of the two classes.

## 3. Results and Discussion

We fine-tuned the Twitter-RoBERTa-base model on our VocalTweets dataset for binary classification, categorizing tweets as Normal or Offensive. Due to the imbalanced nature of the dataset, we observed relatively low performance in the Offensive class, which is the least occurring category. This imbalance contributed to lower F1 scores for the Offensive class. We trained the model across five runs, and the label-wise F1 scores are summarized in table 1 below:

Table 1: Label-wise F1 score after training

| Metric | F1 Score |
|---|---|
| Normal F1 | 91.2 ± 0.4 |
| Offensive F1 | 57.8 ± 3.5 |
| Macro F1 | 74.5 ± 1.9 |

In addition to the overall performance, we examined the model's performance across different language categories. Table 2 below presents the F1 scores for each language:

Table 2: Model performance across the language categories

| Language | Normal F1 | Offensive F1 | Macro F1 |
|---|---|---|---|
| English | 94.1 ± 0.3 | 59.7 ± 0.8 | 76.9 ± 0.5 |
| Code-Switched | 93.8 ± 0.8 | 76.1 ± 3.5 | 85.0 ± 2.1 |
| Naija | 85.7 ± 1.9 | 63.9 ± 1.3 | 74.8 ± 1.4 |
| Yoruba | 85.5 ± 9.2 | 67.3 ± 24.1 | 76.4 ± 16.6 |

The results indicate that the model performs best for the Code-Switched language category, followed closely by the other categories, which exhibit minor differences. While achieving a relatively high Normal F1 score, English has a lower Offensive F1 score due to a larger representation of Offensive sentiments in this category, leading to a lower overall Macro F1 score of 76.9. Table 3 below shows the language distribution of the test set.

Table 3: Language distribution in the test set

| Language | Count | Percentage (%) |
| --- | --- | --- |
| English | 585 | 63.9 |
| Naija | 249 | 27.2 |
| Code-Switched | 71 | 7.8 |
| Yoruba | 11 | 1.2 |

For the cross-corpus experiment, we trained the Twitter-RoBERTa-base model on the OLID dataset and evaluated its performance on both the VocalTweets and OLID datasets. The experiment results using zero-shot transfer are presented in Table 4. We obtained high scores when we trained on OLID and evaluated using the OLID test set. However, testing the model on different corpora yielded lower F1 scores; for instance, OLID → VocalTweets resulted in an F1 score of 69.2. Notably, the OLID dataset has approximately 50% of its data labeled as Offensive, while the VocalTweets dataset contains only 17% of such data. This discrepancy likely contributes to the lower F1 score for the Offensive class.

Table 4: Cross-corpus experimental result across each dataset

| Experiment | Normal F1 | Offensive F1 | Macro F1 |
| --- | --- | --- | --- |
| OLID → OLID | 89.5 ± 1.1 | 79.6 ± 2.6 | 84.5 ± 1.9 |
| VocalTweets → OLID | 72.5 ± 3.7 | 61.1 ± 0.9 | 66.8 ± 2.0 |
| VocalTweets → VocalTweets | 97.2 ± 1.5 | 87.8 ± 6.5 | 92.5 ± 4.0 |
| OLID → VocalTweets | 91.2 ± 0.1 | 47.2 ± 2.3 | 69.2 ± 1.2 |

We evaluated the model's performance further by generating a confusion matrix for one of the trained models. Table 6 displays the results of 10 random classifications performed by the model, highlighting its struggles with the Offensive class; however, it accurately predicted 2 out of the 3 Offensive and misclassified a Normal tweet as Offensive in the sample.

The confusion matrix in Table 5 reveals that Offensive cases are occasionally misclassified as Normal. To enhance the model's ability to predict Offensive tweets accurately, future work should focus on augmenting the dataset with additional Offensive tweets, thereby improving overall model performance.

Table 5: Classification matrix

|  | **Normal** | **Offensive** |
| --- | --- | --- |
| Normal | 697 | 51 |
| Offensive | 66 | 102 |

## 4. Conclusions

We present the VocalTweets dataset for offensive language detection, comprising tweets from 12 famous Nigerian musicians. This dataset is characterized by its code-switched and multilingual nature. We conduct an empirical evaluation using the binary classification method to classify tweets as Normal or Offensive; this results in a Macro F1-Score of 74.5, which shows that the dataset faces a challenge with the Offensive class. To assess the generalization ability of our model, we conducted cross-corpus experiments between the

VocalTweets and OLID datasets, which indicates that our dataset falls slightly short when compared to OLID due to its imbalance nature, the difference is minimal, and it suggests that despite the nature of the dataset we are still able to achieve some level of accuracy compared to OLID which is well balanced. VocalTweets is a valuable resource for advancing research in offensive language detection, particularly for low-resource languages like Yoruba and Nigerian Pidgin.

Future work could focus on augmenting the VocalTweets dataset with additional Offensive tweets to ensure more balanced classes between Normal and Offensive courses to improve the model's performance and robustness. Additionally, cross-corpus experiments could be expanded beyond the OLID dataset to provide more understanding of the model's generalization ability. The dataset can also be expanded to include more Nigerian musicians who are not included in this study and other famous African musicians; this will help create a more diverse dataset that extends beyond Nigerian languages and also other African languages. The dataset can still be expanded to include tweets from famous influencers and public personalities to get a sense of offensive language usage by popular people that influences people's opinions on social media.

# Appendices

## A. Annotation guide for VocalTweets

This guideline gives details on how to annotate tweets within the VocalTweets dataset, focusing on offensive and normal language. As an annotator, it is crucial to approach this task with objectivity and consistency. Annotating offensive content can be psychologically distressing. If you feel anxious or uncomfortable during the annotation process, please take a break or stop the task and seek support.

## Definitions

- **Normal**: Any expression that does not contain offensive language and adheres to societal norms. This includes tweets that are neutral or positive in tone.

**Offensive:** Language that is disrespectful or derogatory towards individuals or groups. This includes insults, derogatory terms, or language intended to harm or belittle others.

Table 6: 10 Random classifications performed by the model

|   | Tweet | Language | True Label | Predicted Label |
|---|---|---|---|---|
| 0 | 😊😊 it's the best accent ever! | English | Normal | Normal |
| 1 | Let the youth embarrass the elders who don't respect themselves. | English | Offensive | Offensive |
| 2 | U don dey post cold, If them send transport now U go use am fill gas Thief! | Naija | Offensive | Offensive |
| 3 | Otilo | Yoruba | Normal | Normal |
| 4 | Live on @fox5ny now tune in!! | English | Normal | Normal |
| 5 | I lied to you sorry | English | Normal | Normal |
| 6 | That's why they must protest with us. | English | Normal | Normal |
| 7 | Congratulations on your win | English | Normal | Normal |
| 8 | Try make money so you fit comot from relationship when you no like | Naija | Normal | Offensive |
| 9 | I never enter gear but shift Abeg. If your soap guy no get Drip, switch shop hehe | English | Offensive | Normal |
| 10 | Hmmm ☐ | English | Normal | Normal |